\documentclass[conference]{IEEEtran}

\usepackage{amsmath,bm}

\def\eqref#1{equation~\ref{#1}}

\def\1{\bm{1}}

\DeclareMathAlphabet{\mathsfit}{\encodingdefault}{\sfdefault}{m}{sl}
\SetMathAlphabet{\mathsfit}{bold}{\encodingdefault}{\sfdefault}{bx}{n}

\usepackage[utf8]{inputenc}
\usepackage[T1]{fontenc}

\usepackage{cite}

\usepackage{amsmath}
\usepackage{amssymb}
\usepackage{amsfonts}
\usepackage{amsthm}
\usepackage{bm}

\usepackage{graphicx}
\usepackage{xcolor}

\usepackage{booktabs}
\usepackage{multirow}
\usepackage{tabularx}
\usepackage{makecell}

\usepackage{pifont}
\usepackage{xspace}
\usepackage{nicefrac}

\usepackage{url}
\usepackage{hyperref}

\usepackage{float}
\usepackage{capt-of}
\usepackage{wrapfig}

\usepackage{algorithm}
\usepackage{algpseudocode}

\usepackage[caption=false,font=footnotesize]{subfig}

\providecommand{\Description}[1]{}

\newcommand{\OURSlong}{%
\underline{B}imodal \underline{E}dge
\underline{E}xplanation via
\underline{P}olarization\xspace}

\newcommand{\OURS}{BEEP\xspace}

\newcommand{\expone}{\ensuremath{h^{(b)}}}
\newcommand{\basemat}{\ensuremath{\hat{\mathbf{S}}^{(b)}}}
\newcommand{\basescore}{\ensuremath{\hat{s}^{(b)}_{uv}}}

\newcommand{\exptwo}{\ensuremath{h^{(g)}}}

\newcommand{\guidedscore}{\ensuremath{\hat{s}^{(g)}_{uv}}}

\newcommand{\pseudoscore}{\ensuremath{s'_{uv}}}
\newcommand{\edgelabel}{\mathbf{S}}

\newcommand{\mypara}[1]{\textbf{#1} }

\newtheorem{lemma}{Lemma}

\newtheorem{corollary}{Corollary}

\begin{document}


\title{Beyond Graph Information Bottleneck:
Creating Bimodal Edge Score Distributions for GNN Explanation}

\author{
\IEEEauthorblockN{Taewook Ham}
\IEEEauthorblockA{
KAIST\\
Daejeon, South Korea\\
htw0618@kaist.ac.kr
}
\and
\IEEEauthorblockN{Cheonwoo Lee}
\IEEEauthorblockA{
KAIST\\
Daejeon, South Korea\\
cheonwoo.lee@kaist.ac.kr
}
\and
\IEEEauthorblockN{Myeong Kong}
\IEEEauthorblockA{
KAIST\\
Daejeon, South Korea\\
myeong.kong@kaist.ac.kr
}
\and
\IEEEauthorblockN{Jaemin Yoo}
\IEEEauthorblockA{
Seoul National University\\
Seoul, South Korea\\
jaeminyoo@snu.ac.kr
}
}

\maketitle


\begin{abstract}
Post-hoc explanation for graph neural networks (GNNs) aims to interpret their predictions by identifying the subgraphs most responsible for a decision. Since discrete subgraphs are not differentiable, existing explainers produce continuous edge-importance scores as output, often yielding blurry, mixed score distributions. The main challenge is the absence of edge-level ground-truth labels, as only the predictions of GNNs are provided as training data. We present BEEP, Bimodal Edge Explanation via Polarization, which converts weak prediction-preserving signals into explicit edge supervision during training. As a result, it markedly improves explanation accuracy while sharply polarizing edges into important versus unimportant ones. We demonstrate that BEEP creates improved shapes of score distributions compared with strong baselines, resulting in more separable explanatory subgraphs. Experiments on four benchmarks show that BEEP improves the accuracy of edge-level explanations over state-of-the-art baselines. To ensure reproducibility, our code is available at \url{https://github.com/Wookido99/BEEP}.
\end{abstract}




\section{Introduction}

Graph neural networks (GNNs) have emerged as a powerful model for learning over graph-structured data~\cite{kipf2016semi, velivckovic2017graph, xu2018powerful, wu2020comprehensive}.
To understand their predictions, post-hoc explanation of GNNs identifies subgraphs responsible for their predictions~\cite{yuan2022explainability}.
Many existing explainers adopt the graph information bottleneck (GIB) principle to learn sparse edge masks that preserve the original prediction~\cite{ying2019gnnexplainer, luo2020parameterized, wang2021towards, zhang2023mixupexplainer, wu2020graph}.
Without ground-truth edge-level supervision, these methods learn continuous edge-importance scores through prediction-preserving objectives, with mask entropy regularization encouraging scores toward 0 or 1.

\begin{figure}[t]
    \centering

    \includegraphics[width=0.72\linewidth]{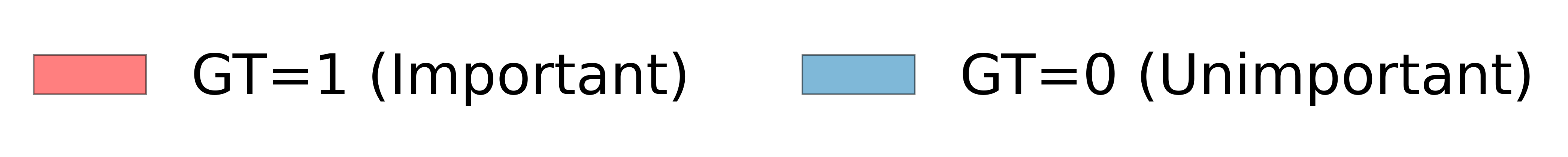}
    \vspace{-10pt}
    
    \makebox[\linewidth][c]{%
        \subfloat[GIB-based.\label{figure:gib_based}]{%
            \includegraphics[width=0.32\linewidth]{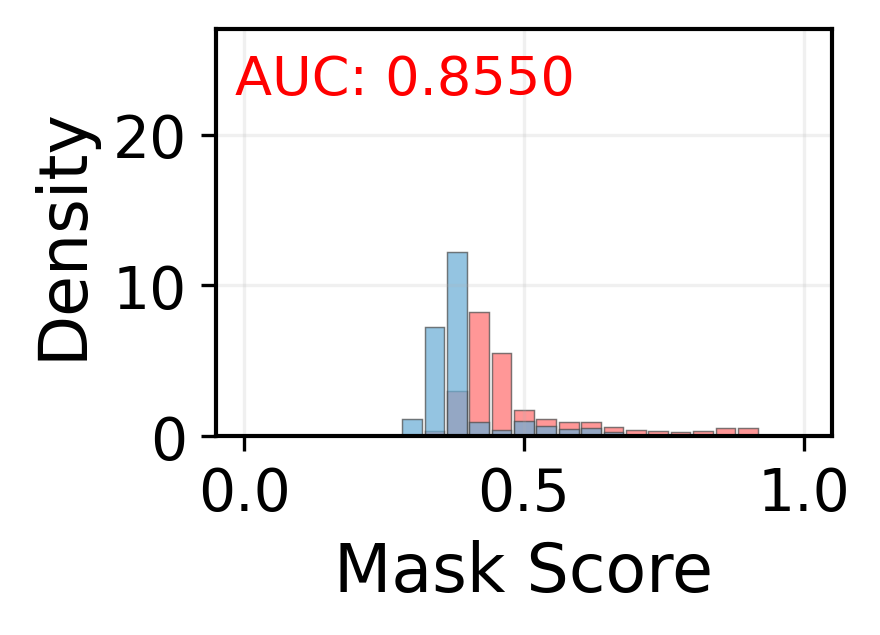}%
        }%
        \hfill
        \subfloat[Over-regularized.\label{figure:over_regularized}]{%
            \includegraphics[width=0.32\linewidth]{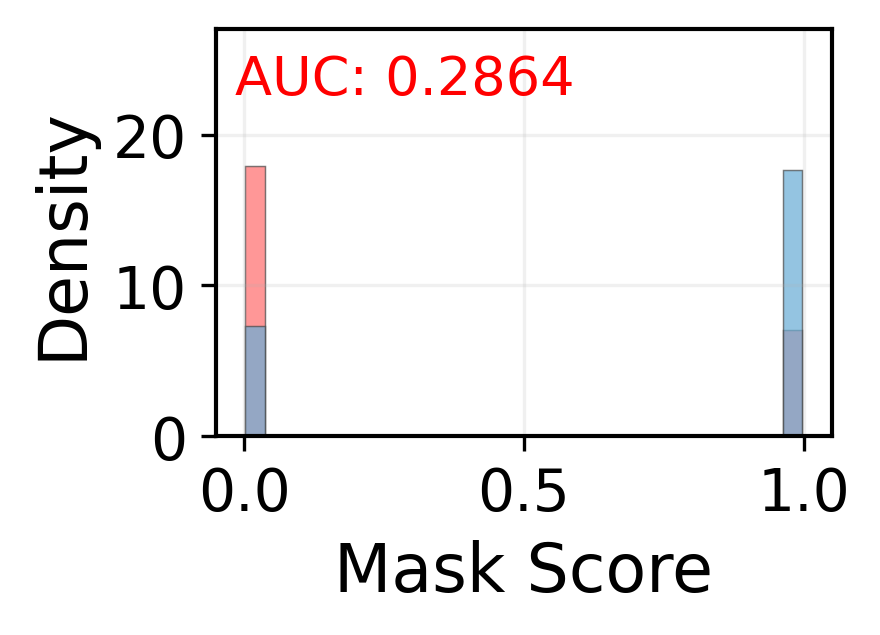}%
        }%
        \hfill
        \subfloat[Our method.\label{figure:our_method}]{%
            \includegraphics[width=0.32\linewidth]{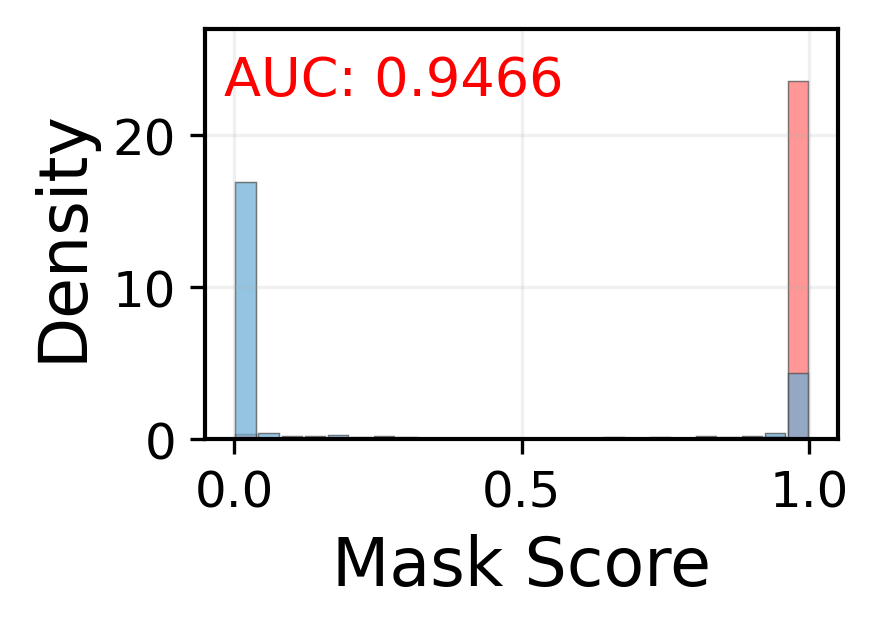}%
        }%
    }

    \caption{
    Edge-score distributions on BA-3motifs.
    (a) GIB-based explanation with ambiguous scores.
    (b) Over-regularization produces over-confident scores.
    (c) \OURS produces a clearly bimodal distribution with high accuracy.
    }
    \label{figure:dist_comparison}
    \vspace{-10pt}
\end{figure}

However, even when these methods achieve high accuracy in explanation benchmarks, the resulting importance scores often fail to exhibit intuitive binarization.
Figure~\ref{figure:gib_based} shows a case where the edge scores are highly ambiguous and poorly polarized regardless of the AUC score. 
A straightforward solution is to strengthen the entropy regularization to enforce sharper score distributions.
However, as shown in Figure~\ref{figure:over_regularized}, naively increasing the regularization strength does not effectively address the problem and instead leads to over-confident predictions.
Note that the colors in the figure refer to ground-truth information that is unavailable in practical usage.

The essential objective of post-hoc explanation is to extract a discrete explanatory subgraph by typically thresholding edge scores or selecting the top-$k$ edges, not to preserve raw scores.
However, existing methods provide no clear criterion for choosing a threshold, and top-$k$ extraction depends only on relative ranking rather than score separability.
Moreover, the appropriate $k$ is generally unknown because the size of the true explanatory subgraph can vary across graphs.
These limitations highlight that explanation quality should be evaluated not only by ranking accuracy~\cite{amara2022graphframex, fang2023evaluating}, but also by whether score distributions are polarized and binarized to support reliable subgraph extraction.



\begin{figure*}[t]
    \centering
    \includegraphics[width=0.9\linewidth]{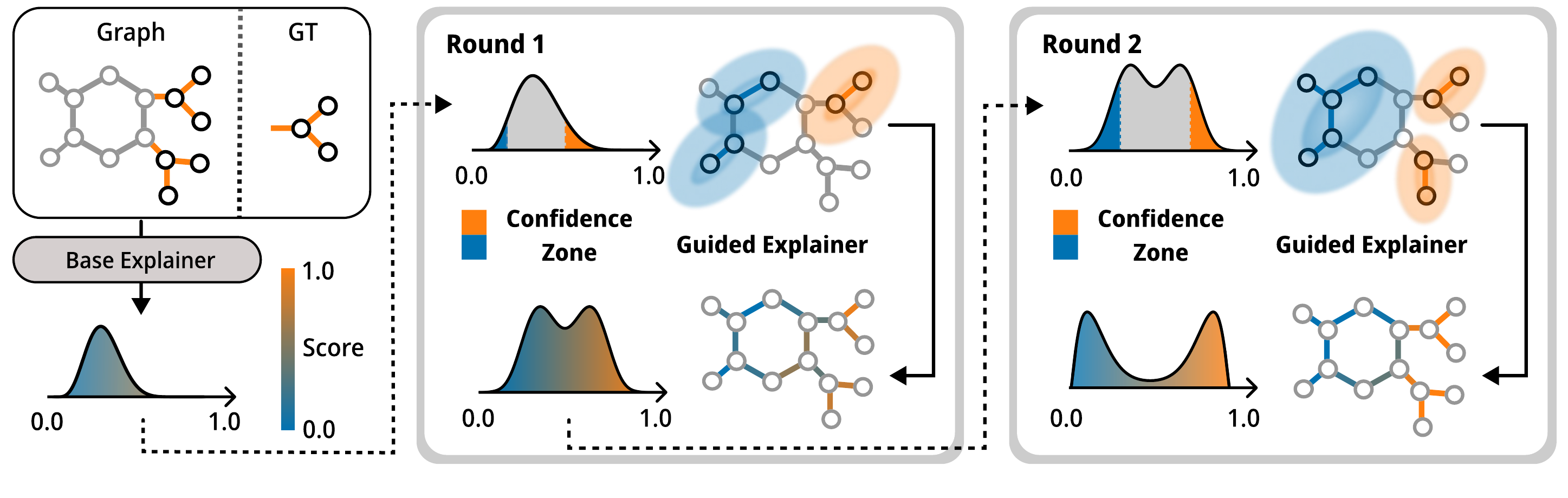}
    \vspace{-15pt}
    \caption{Overall workflow of proposed \OURS.
    Edges falling within the confidence zones (orange or blue regions) are converted into pseudo-labels.
    The guided explainer propagates information from these edges across the graph.
    Subsequent rounds repeat pseudo-labeling and fine-tune the guided explainer.}
    
     \vspace{-15pt}
    \label{fig:stex} 
\end{figure*}
As a solution, we propose \OURS (\OURSlong), which is a general self-guided framework for enhancing the explanation quality of already trained explainers.
\OURS does not require modifying or retraining an existing explainer; it extracts high-confidence signals from the explainer and uses them to train a guided explainer with self-supervision.
Unlike previous methods that rely on unsupervised mutual information, \OURS provides clear guidance to the guided explainer by propagating pseudo-labels derived from the base explainer.
This approach results in a clear, binarized score distribution that separates important and unimportant edges even without complicated regularization.

Our contributions are summarized as follows: (i) \textbf{Extensive Experiments:} We demonstrate improved explanation accuracy and clearer score polarization over baselines on synthetic and real-world datasets, and validate robustness across ablation settings and computational efficiency. (ii) \textbf{Theoretical Analysis:} We analyze how pseudo-label supervision drives scores on confidence-zone edges toward binary targets. (iii) \textbf{Pseudo-labeling Strategy:} We propose a pseudo-labeling strategy that converts soft edge scores into reliable discrete signals, systematically pushing edge scores away from mid-range values.


    

    
\vspace{-3pt}
\section{Preliminaries and Related Work}
\label{Preliminaries}


\mypara{Graph Classification.}
Let \( G = (\mathbf{X}, \mathbf{A}) \) be an undirected graph, where  
\(
\mathbf{X} \in \mathbb{R}^{|\mathcal{V}| \times d}
\) is the node feature matrix,  
\(
\mathbf{A} \in \{0,1\}^{|\mathcal{V}| \times |\mathcal{V}|}
\) is the adjacency matrix,  
\( \mathcal{V} \) is the set of nodes, and \( d \) is the feature size.
For $\mathbf{A}$, each element is $a_{uv}=1$ if nodes $u$ and $v$ are connected and $0$ otherwise. The edge set is $\mathcal{E} \subseteq \mathcal{V} \times \mathcal{V}$, where $(u,v)\in\mathcal{E}$ if and only if $a_{uv}=1$.
In the \emph{graph classification} setting, each graph \( G\) is associated with a label \( Y \in \{1,\dots,C\} \) that describes its global structural or semantic property, where $C$ is the number of classes.
Given a set $\mathcal{G}$ of training graphs, we assume a GNN classifier $f$ trained on $\mathcal{G}$ which is our main target of explanation.

\mypara{Post-hoc Explanation of GNNs.}
Given a trained GNN $f$, \emph{post-hoc explanation} aims to produce a subgraph $G^*$ that most significantly contributes to the GNN’s prediction \( \smash{\hat{Y} = f(G)}\).
Although the goal of explanation is to find a discrete subgraph, most existing explainers return continuous edge importances $\hat{\mathbf{S}} = h(f, G) \in [0, 1]^{|\mathcal{V}| \times |\mathcal{V}|}$, where $h$ denotes an explainer model.
Since the problem is unsupervised, the evaluation of the generated subgraph is typically done by comparing it with the ground-truth edge labels \( \edgelabel \in \{0, 1\}^{|\mathcal{V}| \times |\mathcal{V}|} \), which represent the importance of each edge for the label $Y$ of graph $G$.
Such ground-truth is only for evaluation and is not provided at training time.


\mypara{Graph Information Bottleneck (GIB).}
With the GIB principle~\cite{wu2020graph}, the ideal explanation of GNN $f$ satisfies
\begin{equation}
  \max_{G^{\ast}} \mathrm{MI}(\hat{Y}, G^{\ast})
  = \max_{G^{\ast}} \big(H(\hat{Y}) - H(\hat{Y} \mid G^{\ast})\big),
\end{equation}
where $H$ is the entropy function.
Since \(H(\hat{Y})\) is fixed by the given $f$, maximizing $\mathrm{MI}(\hat{Y}, G^{\ast})$ is equivalent to minimizing \(\smash{H(\hat{Y} \mid G^*)}\), which makes the extracted subgraph $G^*$ have the same class information as in the original graph $G$.
To enforce the compactness of explanation with sparse $G^*$, prior works typically include a regularizer term for its size~\cite{luo2020parameterized, wang2021towards, zhang2023mixupexplainer, zhang2024regexplainer, chen2024generating}.
This constraint encourages the explainer to focus on a minimal set of salient edges.

\mypara{Related Work.}
The principle of explanation through GIB has been consistently adopted across a wide spectrum of works, from early graph mask learning in GNNExplainer~\cite{ying2019gnnexplainer} to parametric prediction with PGExplainer~\cite{luo2020parameterized} and fine-tuning strategies in ReFine~\cite{wang2021towards}. 
To address distribution shifts and improve generalization, later GIB-based methods such as MixupExplainer~\cite{zhang2023mixupexplainer} and ProxyExplainer~\cite{chen2024generating} introduced graph mixing and generation.





\section{Proposed Method: \OURS}
\label{method}

\subsection{Pseudo-labeling via Confidence Zone}
\label{ssec:pseudo}

Given a graph, \OURS obtains continuous mask scores from the base explainer, which may be inaccurate~\cite{lee2013pseudo}.
To acquire confident pseudo-labels for self-guidance, we use two thresholds determined from the score distribution.
Edges above the upper threshold and below the lower threshold form the positive and negative \emph{confidence zones}, denoted by $L_+$ and $L_-$, and receive positive and negative pseudo-labels, respectively.
Edges between the thresholds form the \emph{gray zone} and are ignored when calculating the loss to prevent over-confidence from potentially noisy labels.

Consider the score distribution on the whole dataset from the base explainer $\expone$, i.e.,
$P_s = \{\!\{\, s \mid s \in \expone(G),\; G \in \mathcal{G} \,\}\!\}$,
where $\{\!\{ \cdot \}\!\}$ denotes a multi-set.
Here, $\expone(G)$ outputs a continuous score matrix, and all its individual entries $s$ collectively form the distribution.  
We then compute the skewness $g_1$~\cite{joanes1998comparing} as $g_1 = \mathbb{E}_{s \sim P_s}\big[(s-\mu)^3\big] / \sigma^3$, where $\mu = \mathbb{E}_s[s]$ and $\sigma^2 = \mathbb{E}_s[(s-\mu)^2]$.
The positive skewness $g_1$ indicates a longer upper tail $L_+$ (positive zone), while the negative value indicates a longer lower tail $L_-$ (negative zone).

We then determine asymmetric quantile levels to minimize noise and improve label quality, using a base cutoff $\alpha_0 \in (0,0.5)$ and a skewness adjustment factor $c>0$: $\alpha_{\text{pos}} = \alpha_0 \bigl(1 + c \cdot \max(g_1,0)\bigr)$ and $\alpha_{\text{neg}} = \alpha_0 \bigl(1 + c \cdot \max(-g_1,0)\bigr)$.
Finally, the lower and upper thresholds are defined as $t_L = Q_{\alpha_{\text{neg}}}(P_s)$ and $t_U = Q_{1-\alpha_{\text{pos}}}(P_s)$, where $Q_{\beta}(P_s)$ denotes the $\beta$-th quantile of the distribution $P_s$.
Edges with scores below the lower threshold $t_L$ are assigned label $0$, those above the upper threshold $t_U$ are assigned label $1$, and edges with scores in between remain unlabeled.


\begin{wrapfigure}{l}{0.48\columnwidth}
    \centering
    \includegraphics[width=\linewidth]{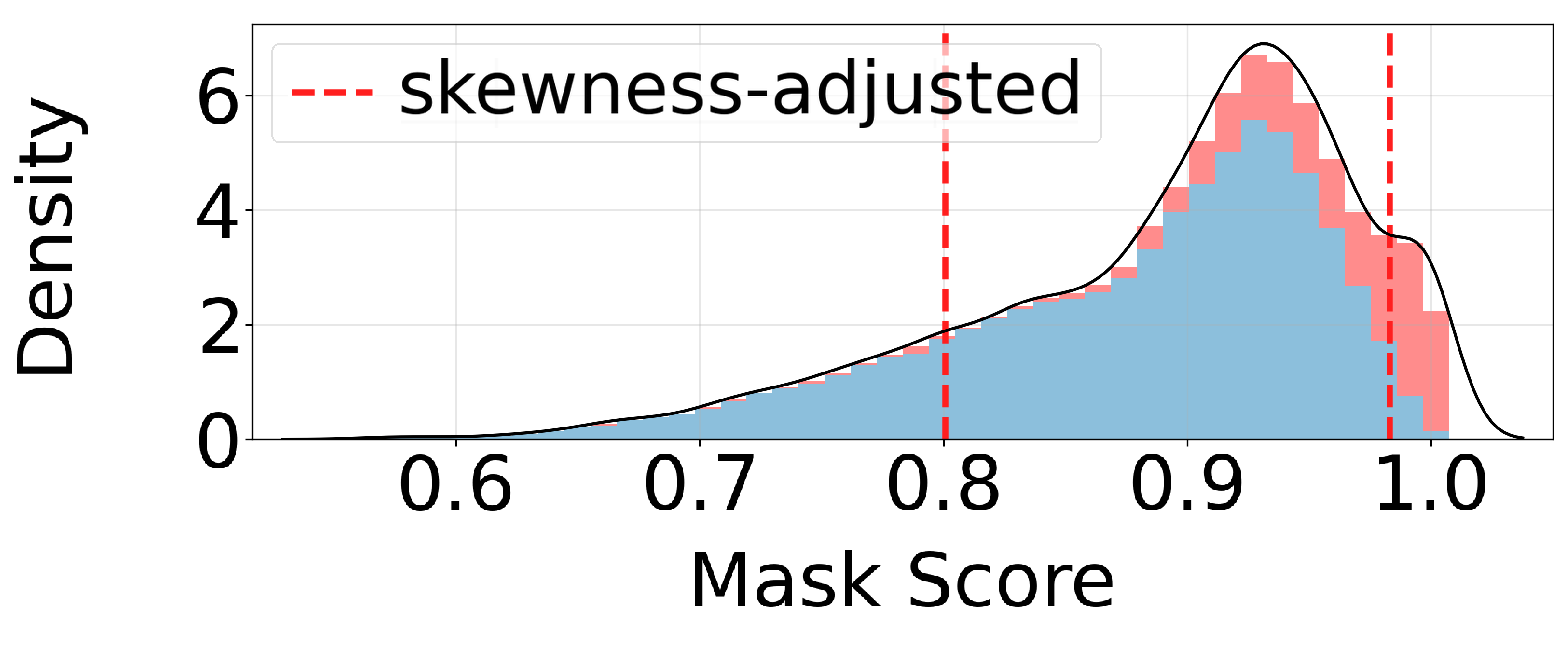}
    \vspace{-15pt}
    \caption{Thresholding on Mutagenicity.
    Red and blue indicate the proportions of ground-truth labels 1 and 0 within each interval, respectively.}
    \label{figure:interesting}
\end{wrapfigure}
One may determine thresholds using fixed symmetric quantiles or a simple mean-based rule. However, these approaches do not explicitly adjust tail proportions for distributional asymmetry and may therefore select noisy pseudo-labels. In Figure~\ref{figure:interesting}, our skewness-adjusted strategy accounts for this asymmetry, yielding more accurate pseudo-labels and more balanced error compositions.

\vspace{-5pt}
\subsection{Training the Guided Explainer}
\label{ssec:training}



The pseudo-labeling creates a new dataset $\mathcal{G}' = \{(G, \mathbf{S}', \mathbf{M})\}$, where $\mathbf{S}' \in \{0, 1\}^{|\mathcal{V}| \times |\mathcal{V}|}$ denotes the pseudo-labels and $\mathbf{M} \in \{0, 1\}^{|\mathcal{V}| \times |\mathcal{V}|}$ represents the supervision mask; $m_{uv} = 1$ means that the pseudo-label is given for edge $(u, v)$.
Our goal is to use the pseudo-labels as stable guidance for training a guided explainer $ \exptwo $.


\mypara{Model Architecture.}
Most explainer models, including all baselines used in our experiments, use an MLP backbone.
This is natural because the explainer takes as input the node embeddings $\mathbf{Z}$ produced by a GNN $f$; a lightweight architecture is sufficient, as the graph structure is already encoded in $\mathbf{Z}$.
In contrast, $\exptwo$ is trained from the pseudo-labels and does not depend on $\mathbf{Z}$.
We therefore employ a GNN backbone for $\exptwo$, allowing the semi-supervised signals to propagate effectively across the graph.

By default, we use a two-layer GCN followed by an edge classifier as $\exptwo$.
Given $G = (\mathbf{X}, \mathbf{A})$, the node embeddings $\mathbf{H}$ are computed as $ \mathbf{H} = \operatorname{ReLU}( \hat{\mathbf{A}} \, \operatorname{ReLU}( \hat{\mathbf{A}} \mathbf{X} \mathbf{W}_{1} ) \mathbf{W}_{2} ) $,  
where 
$ \hat{\mathbf{A}} \in \mathbb{R}^{|\mathcal{V}| \times |\mathcal{V}|} $ is the symmetrically normalized adjacency matrix, and
$ \mathbf{W}_{1} $ and $ \mathbf{W}_{2} $ are learnable weights.
For each edge $(u,v)$,
we extract the concatenated node embeddings and pass them through a linear edge classifier to obtain the final edge importance,
$\guidedscore = \sigma([\mathbf{h}_u \Vert \mathbf{h}_v]\mathbf{W}_3 + \mathbf{b})$,
where $\mathbf{W}_3$ and $\mathbf{b}$ are learnable parameters, and $\sigma(\cdot)$ is the sigmoid function.

\mypara{Training.}
The fraction of important edges varies across datasets, with many edges being easy negatives.
We thus adopt a \emph{positive class-weighted} binary cross-entropy (BCE) loss to account for this imbalance:
\begin{equation}
\label{eq:bce-loss-haha}
\resizebox{0.9\columnwidth}{!}{$
\mathcal{L}_{g}(\exptwo)
= \sum_{(u,v)\in\mathcal{E}}
w^{+} s'_{uv}\log\guidedscore
+ (1-s'_{uv})\log(1-\guidedscore)
$}
\end{equation}
where $\mathcal{E}$ is the set of edges in $G$, $s'_{uv} \in \{0,1\}$ is the pseudo-label for edge $(u,v)$, and $w^{+}$ is the weight assigned to positive edges to compensate for their scarcity.
\begin{table*}[t]
\centering

\caption{
Performance comparison between base explainers and \OURS across different datasets in terms of four evaluation metrics.
}
\label{table:main_table}

\footnotesize
\setlength{\tabcolsep}{2.5pt}
\renewcommand{\arraystretch}{1.0}
\vspace{-5pt}
\resizebox{\textwidth}{!}{%
\begin{tabular}{l | c | cc | cc | cc | cc}
\toprule

\multirow{2}{*}{\textbf{Base Explainer}}
& \multirow{2}{*}{\textbf{Dataset}}
& \multicolumn{2}{c}{\textbf{AUC}}
& \multicolumn{2}{c}{\textbf{Binarization score}}
& \multicolumn{2}{c}{\textbf{Bimodality}}
& \multicolumn{2}{c}{\textbf{F1}}
\\

\cmidrule(lr){3-4}
\cmidrule(lr){5-6}
\cmidrule(lr){7-8}
\cmidrule(lr){9-10}

&
& Base & Guided
& Base & Guided
& Base & Guided
& Base & Guided
\\

\midrule

\multirow{4}{*}{PGExplainer}
 & BA-3motifs
 & $0.930 \pm 0.025$ & $\mathbf{0.936 \pm 0.031}$
 & $0.934 \pm 0.028$ & $\mathbf{0.940 \pm 0.026}$
 & $\mathbf{0.954 \pm 0.022}$ & $0.949 \pm 0.023$
 & $0.882 \pm 0.013$ & $\mathbf{0.899 \pm 0.018}$ \\

 & Mutagenicity
 & $0.855 \pm 0.070$ & $\mathbf{0.968 \pm 0.028}$
 & $0.245 \pm 0.144$ & $\mathbf{0.426 \pm 0.287}$
 & $0.314 \pm 0.035$ & $\mathbf{0.531 \pm 0.233}$
 & $0.278 \pm 0.005$ & $\mathbf{0.376 \pm 0.066}$ \\

 & MNIST-75sp
 & $0.729 \pm 0.001$ & $\mathbf{0.876 \pm 0.074}$
 & $0.172 \pm 0.330$ & $\mathbf{0.742 \pm 0.154}$
 & $0.157 \pm 0.053$ & $\mathbf{0.709 \pm 0.067}$
 & $0.498 \pm 0.001$ & $\mathbf{0.634 \pm 0.077}$ \\

 & Fluoride-Carbonyl
 & $0.853 \pm 0.004$ & $\mathbf{0.895 \pm 0.011}$
 & $0.000 \pm 0.000$ & $\mathbf{0.799 \pm 0.094}$
 & $0.101 \pm 0.079$ & $\mathbf{0.614 \pm 0.180}$
 & $0.562 \pm 0.005$ & $\mathbf{0.625 \pm 0.022}$ \\

\midrule

\multirow{4}{*}{GSAT}
 & BA-3motifs
 & $0.937 \pm 0.008$ & $\mathbf{0.957 \pm 0.003}$
 & $0.540 \pm 0.070$ & $\mathbf{0.956 \pm 0.006}$
 & $0.775 \pm 0.090$ & $\mathbf{0.966 \pm 0.005}$
 & $0.903 \pm 0.009$ & $\mathbf{0.925 \pm 0.003}$ \\

 & Mutagenicity
 & $0.853 \pm 0.037$ & $\mathbf{0.993 \pm 0.006}$
 & $0.065 \pm 0.027$ & $\mathbf{0.571 \pm 0.323}$
 & $0.162 \pm 0.089$ & $\mathbf{0.714 \pm 0.173}$
 & $0.322 \pm 0.018$ & $\mathbf{0.443 \pm 0.056}$ \\

 & MNIST-75sp
 & $0.730 \pm 0.014$ & $\mathbf{0.890 \pm 0.020}$
 & $0.178 \pm 0.050$ & $\mathbf{0.754 \pm 0.166}$
 & $0.182 \pm 0.048$ & $\mathbf{0.787 \pm 0.129}$
 & $0.526 \pm 0.008$ & $\mathbf{0.675 \pm 0.023}$ \\

 & Fluoride-Carbonyl
 & $0.846 \pm 0.018$ & $\mathbf{0.882 \pm 0.017}$
 & $0.807 \pm 0.096$ & $\mathbf{0.973 \pm 0.004}$
 & $0.398 \pm 0.061$ & $\mathbf{0.979 \pm 0.002}$
 & $0.455 \pm 0.013$ & $\mathbf{0.624 \pm 0.039}$ \\

\midrule

\multirow{4}{*}{MixupExplainer}
 & BA-3motifs
 & $\mathbf{0.965 \pm 0.002}$ & $0.961 \pm 0.003$
 & $0.766 \pm 0.086$ & $\mathbf{0.928 \pm 0.008}$
 & $0.788 \pm 0.075$ & $\mathbf{0.937 \pm 0.009}$
 & $\mathbf{0.932 \pm 0.008}$ & $0.919 \pm 0.012$ \\

 & Mutagenicity
 & $0.843 \pm 0.016$ & $\mathbf{0.974 \pm 0.026}$
 & $0.058 \pm 0.043$ & $\mathbf{0.591 \pm 0.355}$
 & $0.453 \pm 0.011$ & $\mathbf{0.524 \pm 0.227}$
 & $0.259 \pm 0.003$ & $\mathbf{0.389 \pm 0.050}$ \\

 & MNIST-75sp
 & $0.741 \pm 0.001$ & $\mathbf{0.900 \pm 0.007}$
 & $0.110 \pm 0.009$ & $\mathbf{0.839 \pm 0.037}$
 & $0.130 \pm 0.003$ & $\mathbf{0.847 \pm 0.033}$
 & $0.506 \pm 0.001$ & $\mathbf{0.664 \pm 0.010}$ \\

 & Fluoride-Carbonyl
 & $\mathbf{0.855 \pm 0.002}$ & $0.838 \pm 0.021$
 & $0.846 \pm 0.020$ & $\mathbf{0.848 \pm 0.055}$
 & $0.883 \pm 0.010$ & $\mathbf{0.922 \pm 0.026}$
 & $\mathbf{0.653 \pm 0.012}$ & $0.556 \pm 0.023$ \\

\midrule

\multirow{4}{*}{ProxyExplainer}
 & BA-3motifs
 & $0.814 \pm 0.036$ & $\mathbf{0.948 \pm 0.009}$
 & $0.690 \pm 0.022$ & $\mathbf{0.960 \pm 0.010}$
 & $0.674 \pm 0.037$ & $\mathbf{0.968 \pm 0.008}$
 & $0.597 \pm 0.017$ & $\mathbf{0.925 \pm 0.003}$ \\

 & Mutagenicity
 & $0.835 \pm 0.078$ & $\mathbf{0.991 \pm 0.008}$
 & $0.011 \pm 0.012$ & $\mathbf{0.560 \pm 0.351}$
 & $0.463 \pm 0.040$ & $\mathbf{0.564 \pm 0.274}$
 & $0.260 \pm 0.002$ & $\mathbf{0.443 \pm 0.142}$ \\

 & MNIST-75sp
 & $0.729 \pm 0.001$ & $\mathbf{0.878 \pm 0.070}$
 & $0.186 \pm 0.386$ & $\mathbf{0.772 \pm 0.146}$
 & $0.175 \pm 0.094$ & $\mathbf{0.734 \pm 0.068}$
 & $0.498 \pm 0.001$ & $\mathbf{0.634 \pm 0.082}$ \\

 & Fluoride-Carbonyl
 & $0.802 \pm 0.001$ & $\mathbf{0.845 \pm 0.014}$
 & $0.896 \pm 0.007$ & $\mathbf{0.899 \pm 0.048}$
 & $0.690 \pm 0.005$ & $\mathbf{0.947 \pm 0.044}$
 & $\mathbf{0.569 \pm 0.001}$ & $0.567 \pm 0.013$ \\

\midrule

\multirow{4}{*}{ConfExplainer}
 & BA-3motifs
 & $0.831 \pm 0.038$ & $\mathbf{0.902 \pm 0.018}$
 & $0.632 \pm 0.330$ & $\mathbf{0.845 \pm 0.083}$
 & $0.764 \pm 0.144$ & $\mathbf{0.884 \pm 0.050}$
 & $0.481 \pm 0.048$ & $\mathbf{0.899 \pm 0.014}$ \\

 & Mutagenicity
 & $0.882 \pm 0.082$ & $\mathbf{0.947 \pm 0.064}$
 & $0.366 \pm 0.494$ & $\mathbf{0.624 \pm 0.160}$
 & $0.480 \pm 0.229$ & $\mathbf{0.688 \pm 0.129}$
 & $0.271 \pm 0.011$ & $\mathbf{0.430 \pm 0.029}$ \\

 & MNIST-75sp
 & $0.743 \pm 0.020$ & $\mathbf{0.878 \pm 0.049}$
 & $0.213 \pm 0.394$ & $\mathbf{0.824 \pm 0.025}$
 & $0.184 \pm 0.108$ & $\mathbf{0.838 \pm 0.020}$
 & $0.506 \pm 0.011$ & $\mathbf{0.650 \pm 0.071}$ \\

 & Fluoride-Carbonyl
 & $0.741 \pm 0.054$ & $\mathbf{0.895 \pm 0.017}$
 & $0.262 \pm 0.392$ & $\mathbf{0.817 \pm 0.241}$
 & $0.193 \pm 0.244$ & $\mathbf{0.617 \pm 0.122}$
 & $0.458 \pm 0.034$ & $\mathbf{0.614 \pm 0.026}$ \\

\midrule

\multirow{4}{*}{EiG-Search}

 & BA-3motifs

 & $0.625 \pm 0.000$ & $\mathbf{0.910 \pm 0.005}$
 & $0.715 \pm 0.000$ & $\mathbf{0.874 \pm 0.043}$
 & $0.697 \pm 0.000$ & $\mathbf{0.904 \pm 0.031}$
 & $0.598 \pm 0.000$ & $\mathbf{0.909 \pm 0.003}$ \\

 & Mutagenicity

 & $0.504 \pm 0.000$ & $\mathbf{0.953 \pm 0.044}$
 & $0.682 \pm 0.000$ & $\mathbf{0.758 \pm 0.075}$
 & $0.705 \pm 0.000$ & $\mathbf{0.754 \pm 0.048}$
 & $0.157 \pm 0.000$ & $\mathbf{0.495 \pm 0.116}$ \\



 & MNIST-75sp
 & $0.519 \pm 0.000$ & $\mathbf{0.760 \pm 0.072}$
 & $\mathbf{0.736 \pm 0.000}$ & $0.680 \pm 0.323$ 
 & $0.728 \pm 0.000$ & $\mathbf{0.759 \pm 0.252}$
 & $0.306 \pm 0.000$ & $\mathbf{0.496 \pm 0.030}$ \\

 & Fluoride-Carbonyl

 & $0.789 \pm 0.000$ & $\mathbf{0.864 \pm 0.049}$
 & $0.770 \pm 0.000$ & $\mathbf{0.943 \pm 0.016}$
 & $0.779 \pm 0.000$ & $\mathbf{0.951 \pm 0.016}$
 & $0.545 \pm 0.000$ & $\mathbf{0.606 \pm 0.059}$ \\

\midrule

\multirow{4}{*}{GOAt}

 & BA-3motifs

 & $0.771 \pm 0.000$ & $\mathbf{0.919 \pm 0.002}$
 & $0.422 \pm 0.000$ & $\mathbf{0.937 \pm 0.021}$
 & $0.363 \pm 0.000$ & $\mathbf{0.943 \pm 0.022}$
 & $0.683 \pm 0.000$ & $\mathbf{0.890 \pm 0.020}$ \\

 & Mutagenicity

 & $0.663 \pm 0.000$ & $\mathbf{0.984 \pm 0.015}$
 & $0.522 \pm 0.000$ & $\mathbf{0.593 \pm 0.327}$
 & $0.542 \pm 0.000$ & $\mathbf{0.746 \pm 0.174}$
 & $0.214 \pm 0.000$ & $\mathbf{0.505 \pm 0.066}$ \\

 & MNIST-75sp
 & $0.504 \pm 0.000$ & $\mathbf{0.908 \pm 0.006}$
 & $\mathbf{0.804 \pm 0.000}$ & $0.734 \pm 0.091$
 & $0.516 \pm 0.000$ & $\mathbf{0.781 \pm 0.069}$
 & $0.417 \pm 0.000$ & $\mathbf{0.708 \pm 0.011}$ \\

 & Fluoride-Carbonyl

 & $0.623 \pm 0.000$ & $\mathbf{0.762 \pm 0.013}$
 & $0.888 \pm 0.000$ & $\mathbf{0.943 \pm 0.013}$
 & $0.836 \pm 0.000$ & $\mathbf{0.951 \pm 0.012}$
 & $0.415 \pm 0.000$ & $\mathbf{0.514 \pm 0.015}$ \\

\bottomrule
\end{tabular}%
}
\vspace{-20pt}
\end{table*}
\begin{figure*}[t]
    \centering


    \subfloat[BA-3motifs\label{fig:multi_hist_ba3}]{%
        \begin{minipage}[t]{0.49\textwidth}
            \centering
            \includegraphics[width=\linewidth]
            {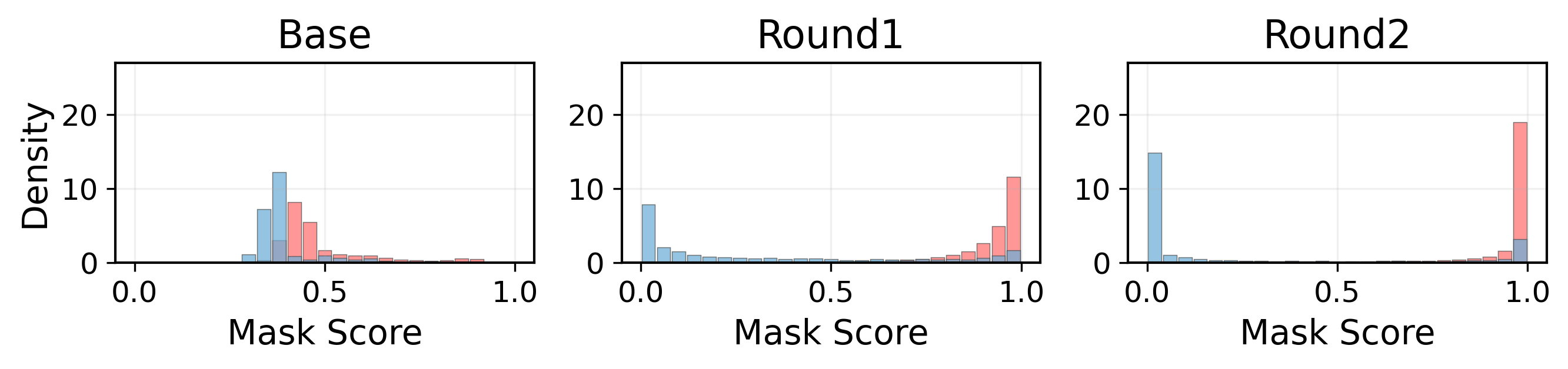}
        \end{minipage}%

    }
    \hfill
    \subfloat[Mutagenicity\label{fig:multi_hist_mutag}]{%
        \begin{minipage}[t]{0.49\textwidth}
            \centering
            \includegraphics[width=\linewidth]
            {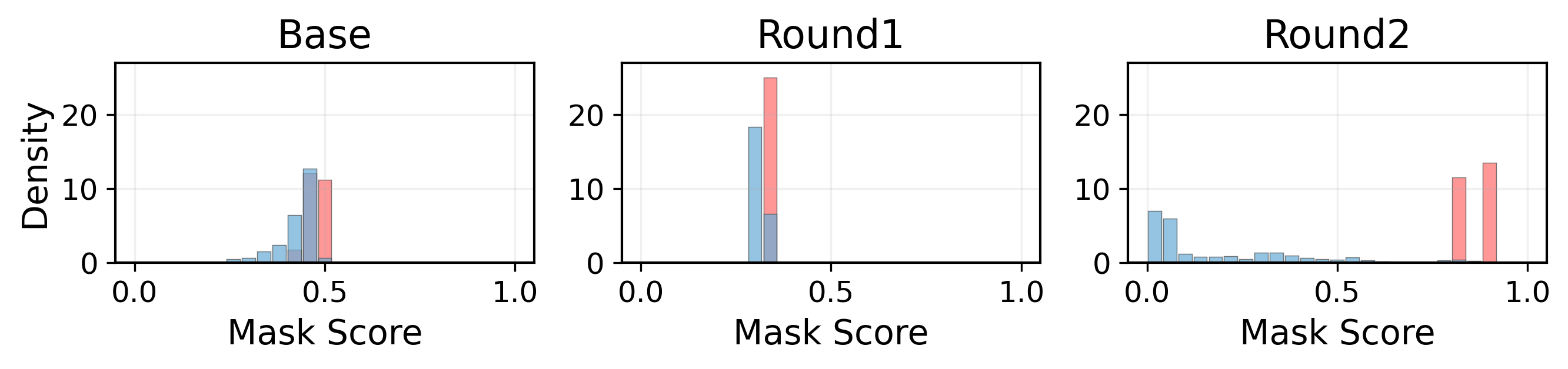}
        \end{minipage}%
    }

    \par\medskip
    \vspace{-15pt}
    \subfloat[Fluoride-Carbonyl\label{fig:multi_hist_fc}]{%
        \begin{minipage}[t]{0.49\textwidth}
            \centering
            \includegraphics[width=\linewidth]
            {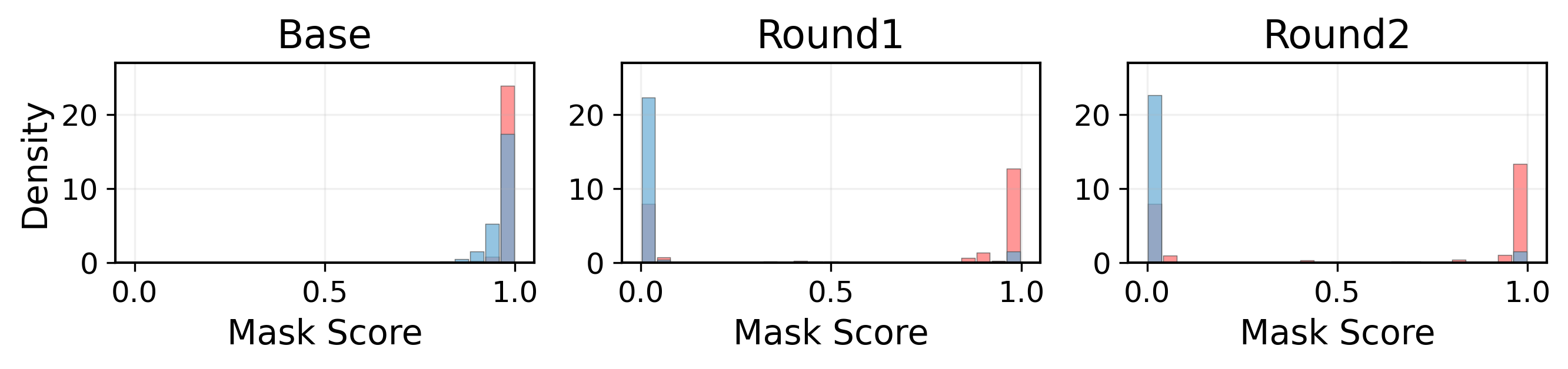}
        \end{minipage}%
    }
    \hfill
    \subfloat[MNIST-75sp\label{fig:multi_hist_mnist}]{%
        \begin{minipage}[t]{0.49\textwidth}
            \centering
            \includegraphics[width=\linewidth]
            {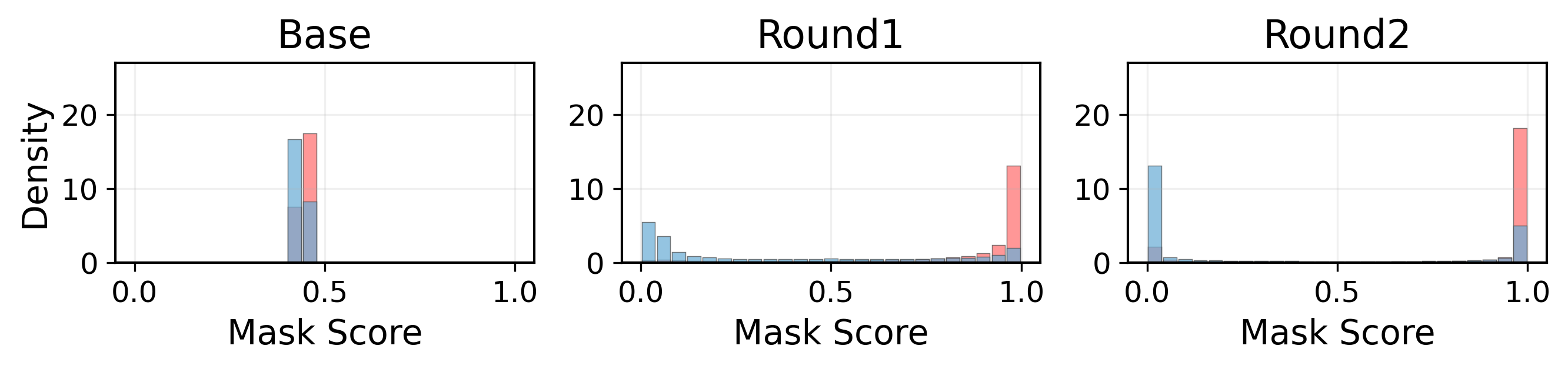}
        \end{minipage}%
    }
    \caption{Score distributions across four datasets.
    For each dataset, the three plots correspond to the base explainer, the first round, and the second round.}
    \label{fig:multi_hist}
    \vspace{-15pt}
\end{figure*}
\vspace{-5pt}
\subsection{Why \OURS Succeeds in Binarization}
\label{ssec:success}

The binarization of edge importance scores might be achieved by adding a suitable regularizer to the GIB objective function, not only by our self-guided framework.
Previous work \cite{luo2020parameterized} proposed an objective function that balances prediction consistency, sparsity, and the binarization of edge importance scores as follows:
\begin{equation}
\label{eq:gib_surrogate}
\resizebox{0.9\columnwidth}{!}{$
\mathcal{L}_{b}(\expone)
= \operatorname{CE}(\hat{Y},\hat{Y}^*)
+ \lambda_1 \sum_{(u,v)\in\mathcal{E}} \basescore
+ \lambda_2 \sum_{(u,v)\in\mathcal{E}} H(\basescore)
$}
\end{equation}
where $H$ is the entropy function minimized if the scalar input is $0$ or $1$, $\hat{Y}^* = f(G^*)$ is the prediction of the GNN $f$ for the extracted subgraph $G^*=(\mathbf X, \basemat)$, and $\basescore\in[0,1]$ is the soft importance score for edge $(u,v)$ produced by the explainer $\expone$.
The first term measures consistency with the original GNN prediction, the second term encourages score sparsity, and the third term promotes binarization, where the coefficients $\lambda_1$ and $\lambda_2$ make a balance.

The entropy penalty is minimized by any binary assignment because $H(0) = H(1) = 0$, irrespective of which edges receive high scores. It therefore encourages extreme scores without distinguishing whether those assignments identify the relevant edges. In the GIB objective, this distinction must be learned through the fidelity term. Consequently, stronger binarization alone does not establish better explanation quality; the relevant goal is to sharpen score separation while retaining useful edge-relevance information.

We now analyze why semi-supervised training drives edge scores toward clear binarization.
To analyze the gradient flow, let $z_{uv}$ denote the pre-sigmoid logit of the guided edge score for edge $(u,v)$, such that $\guidedscore=\sigma(z_{uv})$.
Let $l_{uv}$ denote the per-edge weighted binary cross-entropy loss from Eq.~(\ref{eq:bce-loss-haha}), i.e., $\mathcal{L}_g = \sum_{(u,v)} l_{uv}$.
Importantly, we account for the \emph{tail noise rate} $\varepsilon_\pm$, which represents the proportion of incorrect pseudo-labels present in these confident regions. As long as the pseudo-labeling is better than random guessing ($\varepsilon_\pm < 1/2$), the expected gradient robustly pushes the scores toward the true binary extremes, effectively polarizing the distribution.

\begin{lemma}[Binarization in the confidence zone]
\label{lem:push}
 
Let the gradient of the per-edge loss $l_{uv}$ with respect to the logit $z_{uv}$ be proportional to $\guidedscore - \pseudoscore$. Assume the tail noise rates satisfy $\varepsilon_\pm < 1/2$.
Conditioned on the positive confidence zone (upper tail) $L_+$, the expected gradient is strictly negative, i.e., $\mathbb{E}[\partial l_{uv}/\partial z_{uv} \mid L_+] < 0$, causing the score to polarize toward $\guidedscore \to 1$. 
Conversely, on the negative confidence zone (lower tail) $L_-$, the expected gradient is strictly positive, i.e., $\mathbb{E}[\partial l_{uv}/\partial z_{uv} \mid L_-] > 0$, causing $\guidedscore \to 0$.
\end{lemma}

\begin{proof}[Proof Sketch]
The assigned pseudo-label $\pseudoscore$ is constant within each tail: $1$ for $L_+$ and $0$ for $L_-$. 
For the upper tail $L_+$, the conditional expectation of the BCE gradient is $\mathbb{E}[\guidedscore - 1 \mid L_+] = \mathbb{E}[\guidedscore \mid L_+] - 1$. 
In gradient descent, a negative gradient causes the logit $z_{uv}$ to increase on average, pushing $\guidedscore$ toward $1$. A symmetric argument also applies to the lower tail $L_-$.
\end{proof}


\begin{corollary}[Binarization in the gray zone]\label{cor:propagate}
For an unlabeled edge $(u, v)$ in the gray zone, one gradient step with learning rate $\eta$ gives
\begin{equation}
\textstyle \Delta \guidedscore \approx -\eta 
\sum_{(i,j) \in L} 
\frac{\partial l_{ij}}{\partial \hat{s}_{ij}^{(g)}} 
\langle \nabla_{\theta} \guidedscore, 
\nabla_{\theta} \hat{s}_{ij}^{(g)} \rangle,
\end{equation}
where $L = L_+ \cup L_-$ contains the edges in the confidence zones. If the inner products are nonnegative, the first-order update
increases the score when positive pseudo-labels dominate the
sum in magnitude, and decreases it when negative pseudo-labels
dominate.
\end{corollary}

\begin{proof}[Theoretical Intuition]
First-order Taylor expansion gives

$\Delta\guidedscore \approx 
-\eta \langle \nabla_{\theta} \guidedscore, 
\nabla_{\theta} \mathcal{L}_g \rangle$. 
Substituting
$\nabla_{\theta}\mathcal{L}_g
=\sum_{(i,j)\in L}
(\partial l_{ij}/\partial\hat{s}_{ij}^{(g)})
\nabla_{\theta}\hat{s}_{ij}^{(g)}$
yields the stated expression.
For weighted BCE with $w_+>0$, the loss derivative is negative on $L_+$ and positive on $L_-$, giving the stated directions. When nearby labeled edges have strong positive gradient alignment, they can therefore transmit binarizing signals to gray-zone edges through shared parameters.

\end{proof}


\vspace{-15pt}
\subsection{Linear Time Complexity of \OURS}
\label{ssec:complexity}


\begin{lemma}[Time complexity of \OURS]\label{lemma:cost}
Let $\mathcal{D}=\{G_g=(V_g,E_g)\}$ be the dataset, $N=\sum_g |V_g|$ be the total number of nodes, and $M=\sum_g |E_g|$ be the number of edges.
Assume the guided explainer is an $L_g$-layer message-passing GNN (hidden dimension $h$) with an edge head of per-edge cost $C_e$,
trained for $T$ epochs per round, and quantiles are computed by sorting.
Then, a single round of \OURS over $\mathcal{D}$ costs
\begin{equation}
\mathcal{O} \Big((T+1)\{L_g(Mh+Nh^2)+MC_e\}+M\log M\Big).
\end{equation}
\end{lemma}
A single round consists of re-scoring, pseudo-labeling, and guided
training. Re-scoring requires one forward pass of the $L_g$-layer
GNN and edge scoring, costing
$\mathcal{O}(L_g(Mh+Nh^2)+MC_e)$.
Pseudo-labeling requires sorting the $M$ edge scores to compute
quantiles, costing $\mathcal{O}(M\log M)$.
Training for $T$ epochs costs
$\mathcal{O}(T\{L_g(Mh+Nh^2)+MC_e\})$.
The time complexity of \OURS is linear with respect to graph size.

\section{Experiments}
\label{experiments}





\subsection{Experimental Settings}

\mypara{Baselines.}
We compare \OURS against five GIB-based methods (PGExplainer~\cite{luo2020parameterized}, GSAT~\cite{miao2022interpretable}, MixupExplainer~\cite{zhang2023mixupexplainer}, ProxyExplainer~\cite{chen2024generating}, and ConfExplainer~\cite{zhang2025your}) and two training-free non-GIB methods (EiG-Search~\cite{pmlr-v235-lu24g} and GOAt~\cite{lu2024goat}).

\mypara{Datasets.}
We run experiments on two types of graph datasets: synthetic and real-world graphs. For synthetic graphs, we use the BA-3motifs~\cite{wang2021towards} and MNIST-75sp datasets~\cite{monti2017geometric}. For real-world data, we include two molecular datasets, Mutagenicity~\cite{debnath1991structure} and Fluoride-Carbonyl~\cite{sanchez2020evaluating}.


\mypara{Evaluation Metrics.}
We evaluate explanation accuracy using AUC and F1, and evaluate the shape of the score distributions using bimodality and binarization metrics~\cite{pfister2013good}.
We measure bimodality by
$\frac{3}{2}\left(\frac{\gamma^2+1}{\kappa}-\frac{1}{3}\right)$,
where $\gamma=\mu_3/\mu_2^{3/2}$, $\kappa=\mu_4/\mu_2^2$, and
$\mu_k=\frac{1}{n}\sum_{i=1}^{n}(p_i-\bar{p})^k$.
We further introduce the binarization score, which is defined as
$\text{Binarization score}(p_1,\dots,p_n)
=\frac{1}{n}\sum_{i=1}^{n}(2p_i-1)^2$.


\mypara{Implementation Details.}
We use each baseline as the base explainer for \OURS and report results after the second round of self-guidance.
We report the mean and standard deviation over 5 random seeds for each experiment.
The main hyperparameters ($\alpha_0$, $c$, $w^+$) are selected via grid search.


\vspace{-5pt}
\subsection{Performance Improvement}

Table~\ref{table:main_table} shows that \OURS generally achieves higher AUC and F1 scores than the corresponding GIB-based explainers across synthetic and real-world benchmarks.
For F1 evaluation, we adopt the mean score as the decision threshold without dataset-specific tuning.
Furthermore, \OURS improves both AUC and F1 scores when applied to non-GIB-based explainers.
These results suggest that the proposed self-supervision framework is effective beyond specific explainer objectives and can generalize well to more diverse explainer settings.

\begin{figure}[t]
    \centering

    \subfloat[BA-3motifs\label{fig:noise_ba3}]{%
        \begin{minipage}[t]{\linewidth}
            \centering
            \includegraphics[width=\linewidth]
            {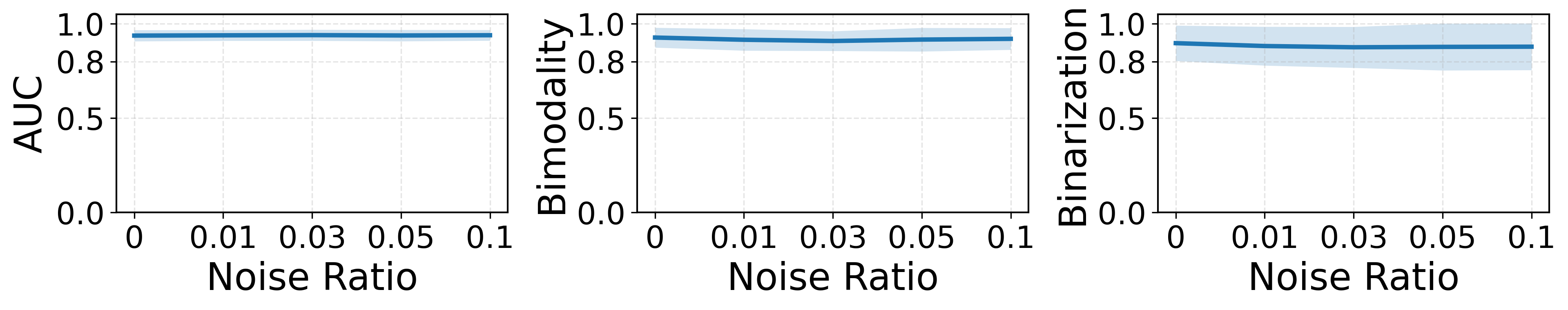}
        \end{minipage}%
    }
    \vspace{-5pt}

    \subfloat[MNIST-75sp\label{fig:noise_mnist}]{%
        \begin{minipage}[t]{\linewidth}
            \centering
            \includegraphics[width=\linewidth]
            {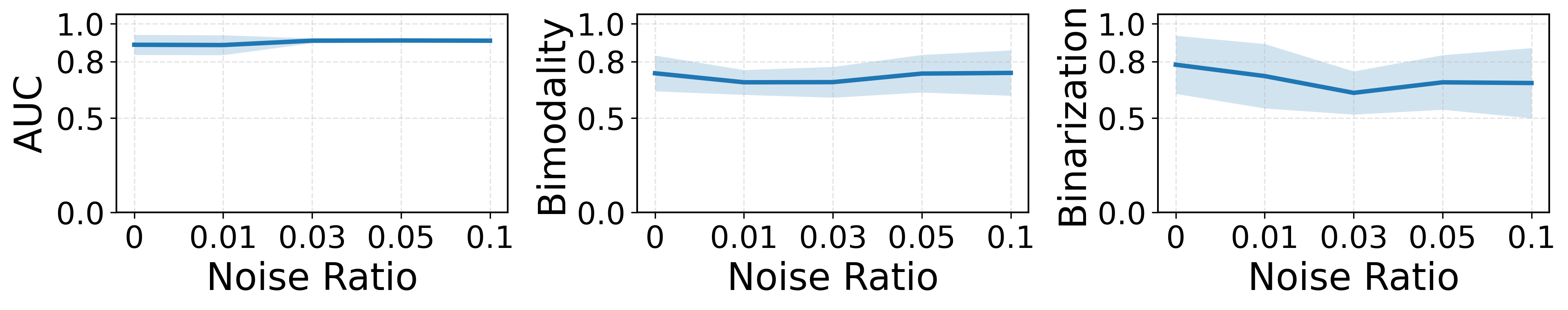}
        \end{minipage}%
    }


    \caption{
    Effect of injected pseudo-label noise on explanation quality.
    }
    \label{fig:noise_analysis}
    \vspace{-15pt}
\end{figure}





\begin{figure}[t]
    \centering

    \includegraphics[width=0.90\linewidth]{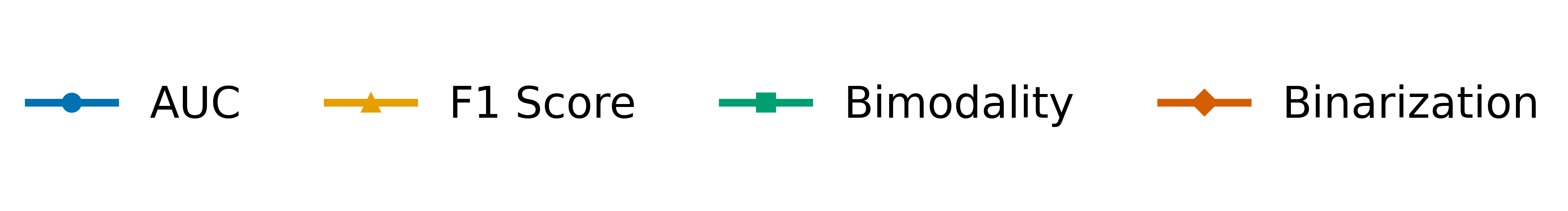}
    \vspace{-5pt}
    
    \includegraphics[width=0.47\linewidth]
    {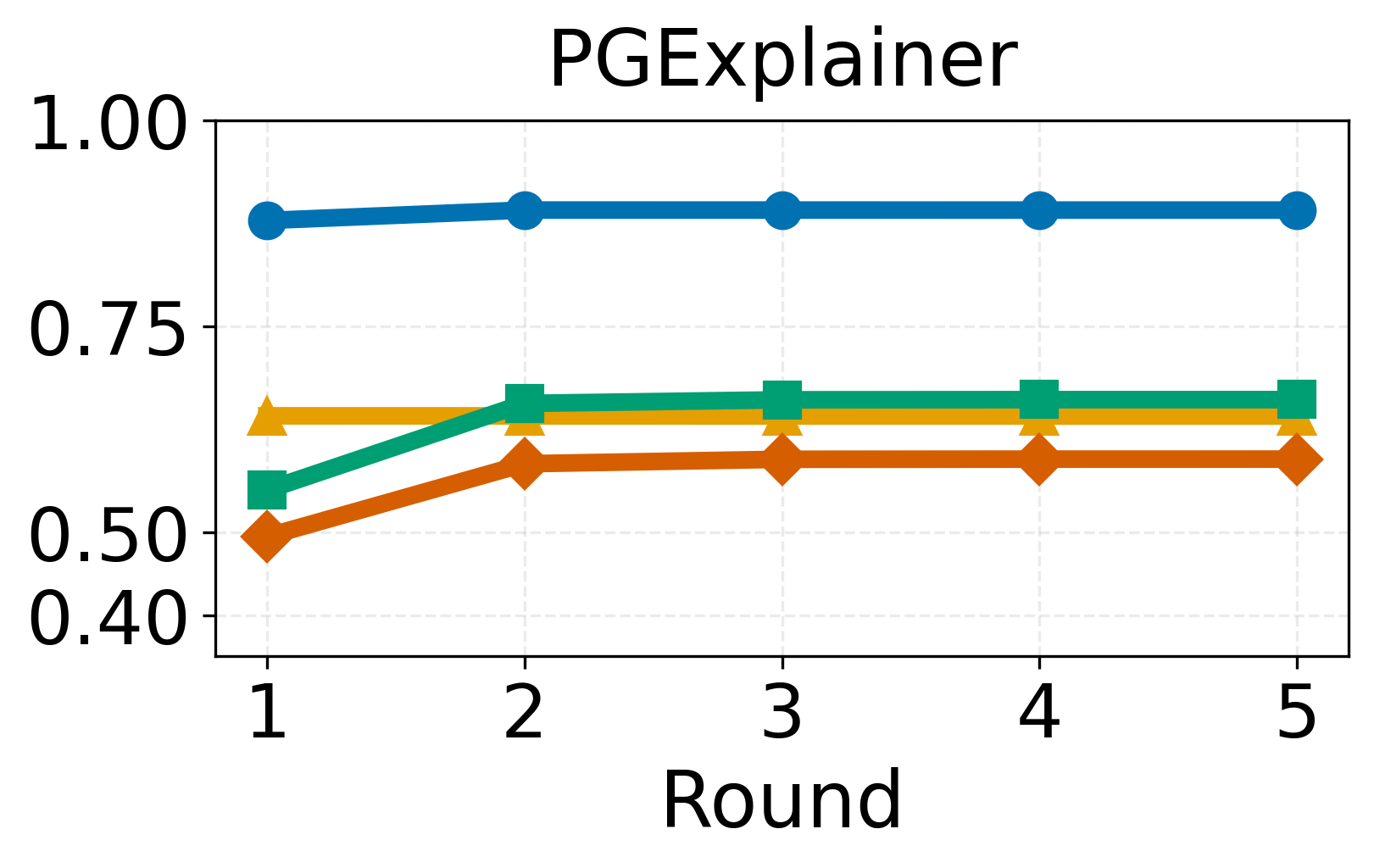}
    \hfill
    \includegraphics[width=0.47\linewidth]
    {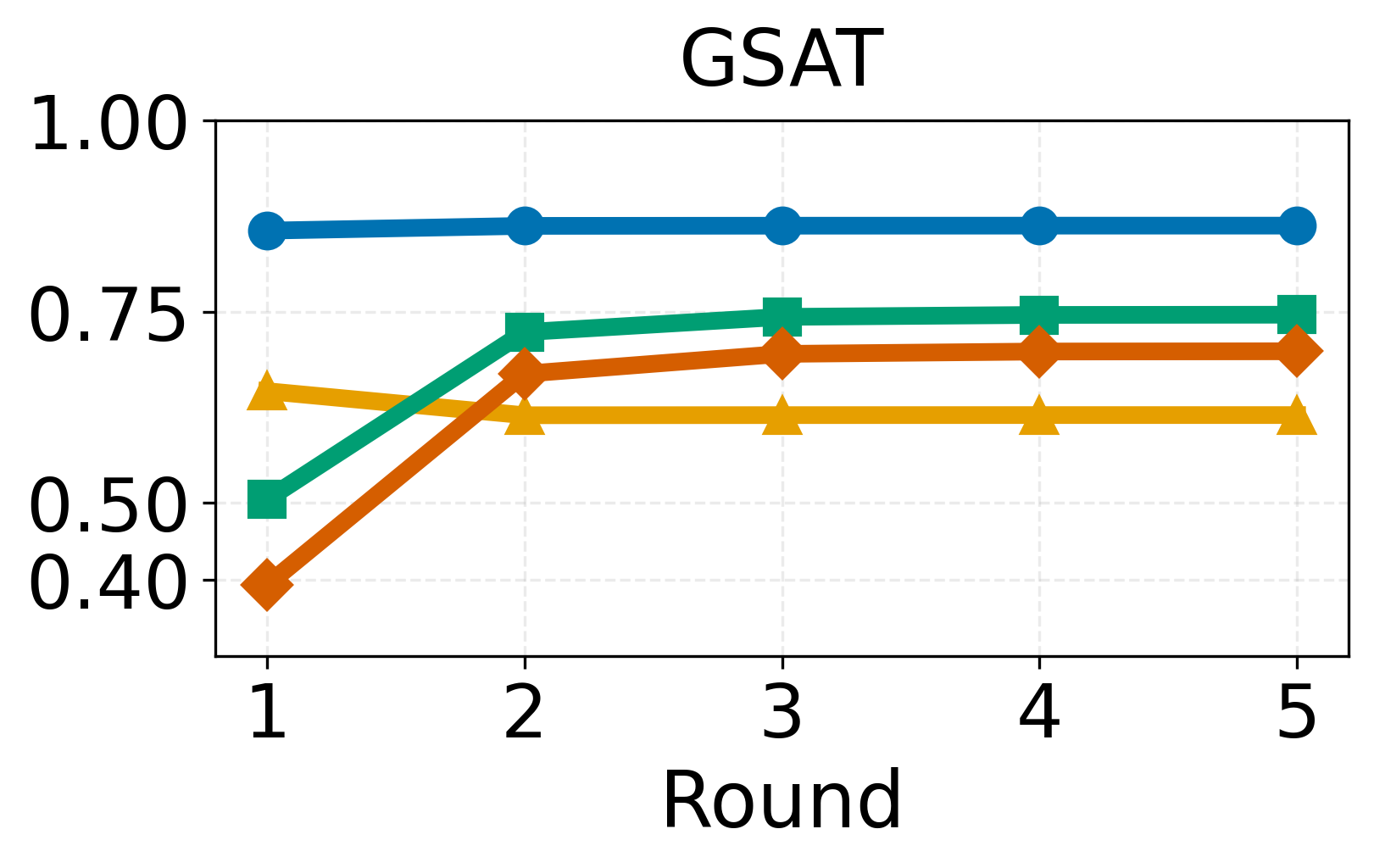}

    \includegraphics[width=0.47\linewidth]
    {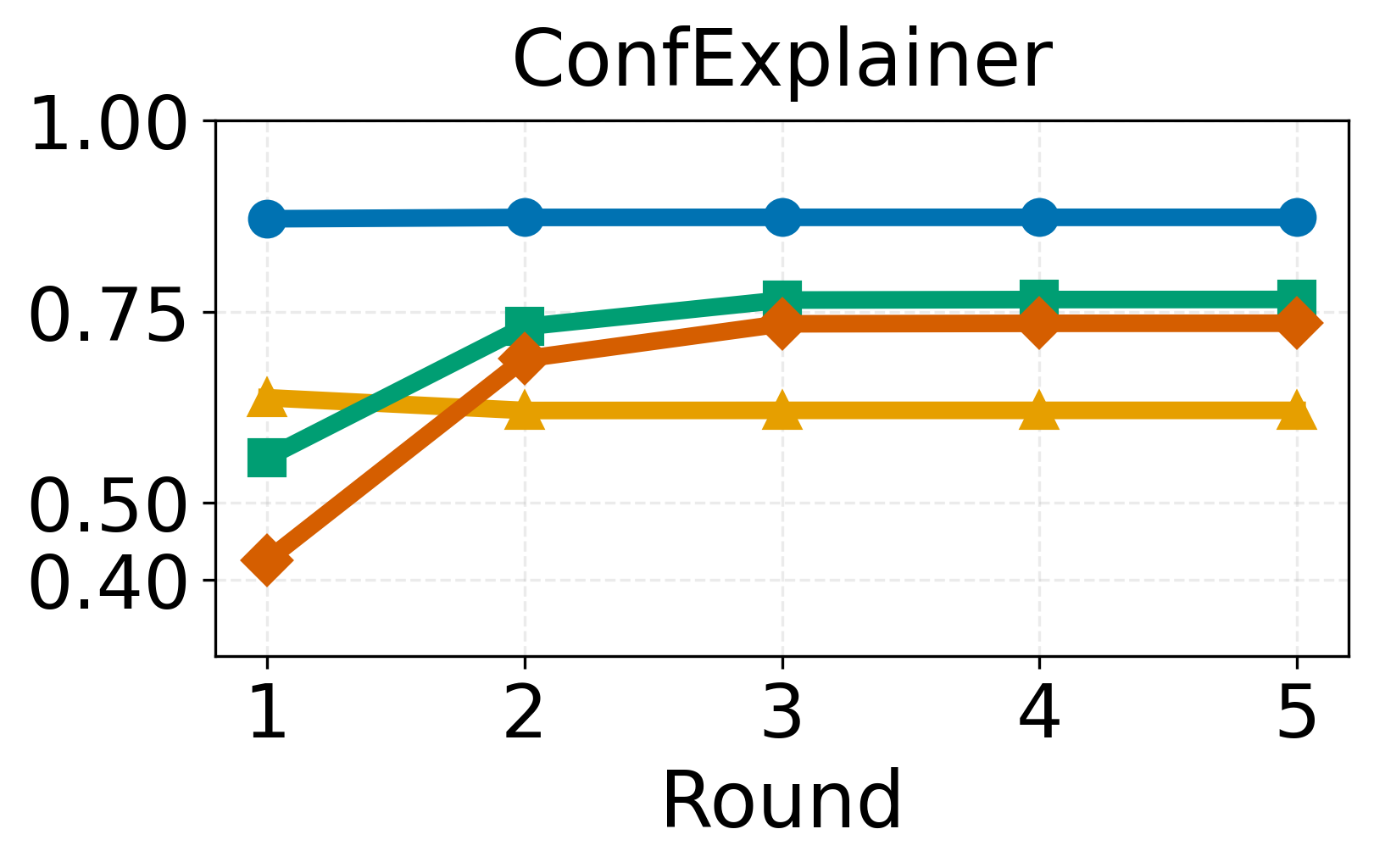}
    \hfill
    \includegraphics[width=0.47\linewidth]
    {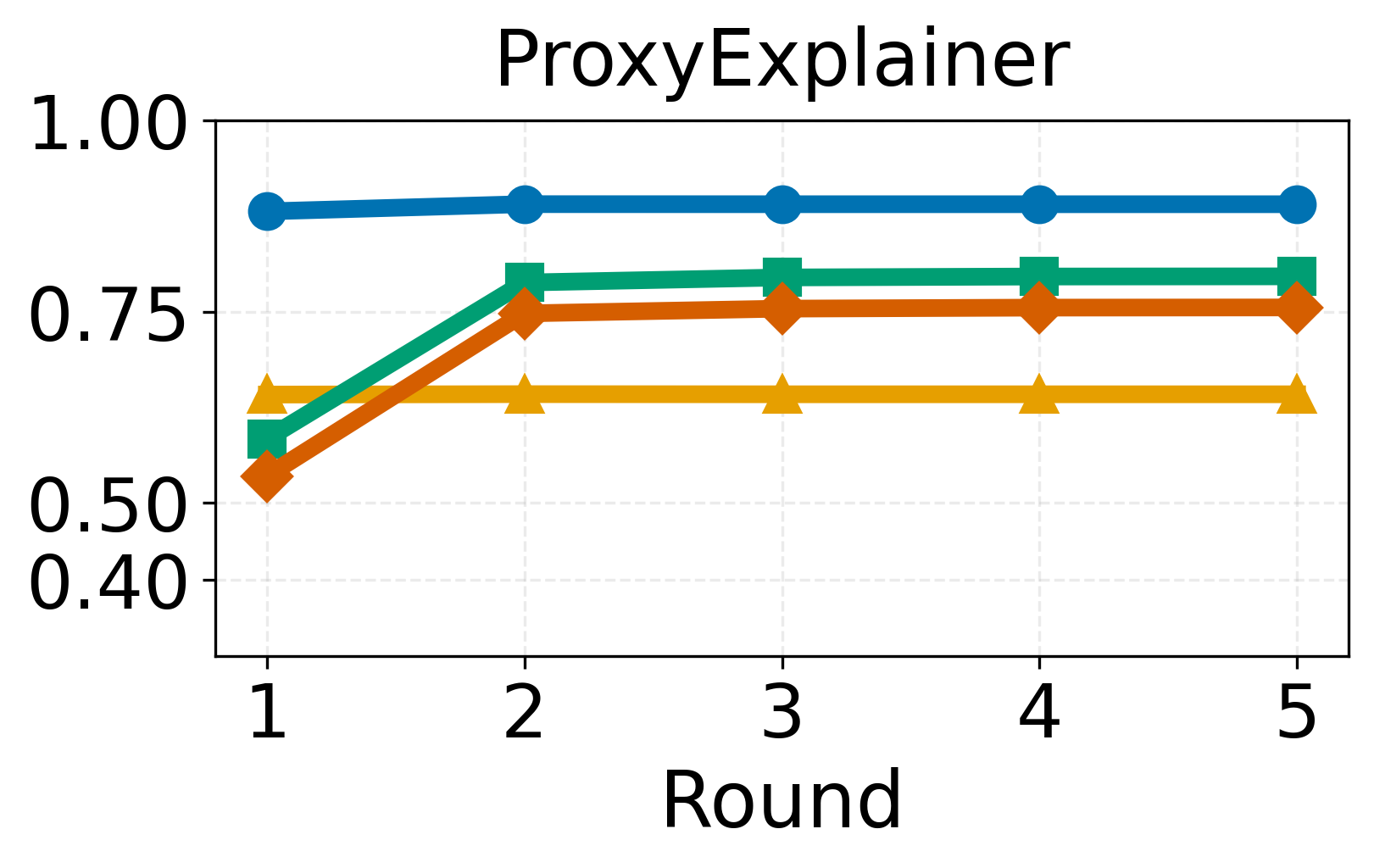}

    \caption{
    Change of various metrics over multiple rounds of \OURS with different base explainers on MNIST-75sp.
    }
    \label{fig:multi}
    \vspace{-10pt}
\end{figure}

\vspace{-5pt}
\subsection{Analysis of Score Distributions}
\label{RQ2}


Figure~\ref{fig:multi_hist} compares the edge-score distributions produced by the base explainer and our guided explainer over self-guided rounds on four datasets.
For the base explainer, the scores are often concentrated in a narrow intermediate region or show substantial overlap between label-$0$ and label-$1$ edges, making it difficult to select a reliable threshold for extracting explanatory subgraphs.
After applying \OURS, label-$0$ edges are increasingly pushed toward low scores, whereas label-$1$ edges are shifted toward high scores.
This trend becomes more pronounced over successive rounds, resulting in progressively more polarized and interpretable score distributions.
\vspace{-5pt}
\begin{table*}[t]
\centering
\caption{Training and inference time (in seconds) across datasets.
Guided training time includes the multi-round training cost (Round 1 + Round 2) and pseudo-label generation time, while guided inference time corresponds to the final Round 2 explainer.}
\vspace{-5pt}
\label{table:time}

\resizebox{\textwidth}{!}{
\begin{tabular}{l|cc|cc|cc|cc||cc|cc|cc|cc}
\toprule

\multirow{3}{*}{\textbf{Base Explainer}}
& \multicolumn{8}{c||}{\textbf{Training Time (s)}}
& \multicolumn{8}{c}{\textbf{Inference Time (s)}} \\

\cmidrule(lr){2-9}
\cmidrule(lr){10-17}

& \multicolumn{2}{c|}{\textbf{BA3}}
& \multicolumn{2}{c|}{\textbf{Mutagenicity}}
& \multicolumn{2}{c|}{\textbf{MNIST}}
& \multicolumn{2}{c||}{\textbf{FC}}

& \multicolumn{2}{c|}{\textbf{BA3}}
& \multicolumn{2}{c|}{\textbf{Mutagenicity}}
& \multicolumn{2}{c|}{\textbf{MNIST}}
& \multicolumn{2}{c}{\textbf{FC}} \\

\cmidrule(lr){2-3}
\cmidrule(lr){4-5}
\cmidrule(lr){6-7}
\cmidrule(lr){8-9}
\cmidrule(lr){10-11}
\cmidrule(lr){12-13}
\cmidrule(lr){14-15}
\cmidrule(lr){16-17}

& \textbf{Base} & \textbf{Guided}
& \textbf{Base} & \textbf{Guided}
& \textbf{Base} & \textbf{Guided}
& \textbf{Base} & \textbf{Guided}

& \textbf{Base} & \textbf{Guided}
& \textbf{Base} & \textbf{Guided}
& \textbf{Base} & \textbf{Guided}
& \textbf{Base} & \textbf{Guided} \\

\midrule

PGExplainer
& \textbf{14.33} & 14.49
& 11.10 & \textbf{5.60}
& 47.69 & \textbf{45.42}
& 45.07 & \textbf{3.76}

& 0.112 & \textbf{0.036}
& 0.054 & \textbf{0.020}
& 0.555 & \textbf{0.446}
& 0.050 & \textbf{0.023} \\

GSAT
& 15.51 & \textbf{14.97}
& 9.12 & \textbf{6.88}
& 69.02 & \textbf{55.74}
& 88.06 & \textbf{6.94}

& 0.086 & \textbf{0.037}
& 0.057 & \textbf{0.020}
& 0.533 & \textbf{0.458}
& 0.055 & \textbf{0.022} \\

MixupExplainer
& 222.20 & \textbf{12.24}
& 124.17 & \textbf{5.91}
& 808.31 & \textbf{56.23}
& 1230.81 & \textbf{3.66}

& 0.114 & \textbf{0.035}
& 0.054 & \textbf{0.022}
& 0.544 & \textbf{0.448}
& 0.049 & \textbf{0.022} \\

ProxyExplainer
& 152.48 & \textbf{17.30}
& 108.76 & \textbf{6.45}
& 859.47 & \textbf{49.63}
& 877.94 & \textbf{3.98}

& 0.120 & \textbf{0.035}
& 0.052 & \textbf{0.021}
& 0.547 & \textbf{0.471}
& 0.048 & \textbf{0.023} \\

ConfExplainer
& 24.00 & \textbf{8.77}
& 29.48 & \textbf{6.97}
& 74.51 & \textbf{58.13}
& 91.17 & \textbf{3.75}

& 0.116 & \textbf{0.038}
& 0.057 & \textbf{0.021}
& 0.549 & \textbf{0.450}
& 0.048 & \textbf{0.022} \\

\bottomrule
\end{tabular}
}
\vspace{-15pt}
\end{table*}

\subsection{Ablation Studies}
\label{ssec:abl-study}

\mypara{Pseudo-label Reliability.}
To evaluate the stability of \OURS under noisy pseudo-labels, we conduct a noise-injection study.
A fixed ratio of pseudo-labels is flipped to the opposite label at each round, while hyperparameters remain identical to those in the main experiments.
In Figure~\ref{fig:noise_analysis}, AUC remains largely unchanged as the noise ratio increases, while bimodality and binarization change only slightly and remain high.

\mypara{Round-wise Improvements.}
In Figure~\ref{fig:multi}, while the AUC shows little change beyond slight improvements, the binarization score and bimodality continue to rise and eventually converge.
For F1 scores, the highest values typically appear in the initial round, after which they decrease slightly but eventually stabilize.

\vspace{-5pt}
\subsection{Robustness to Hyperparameter Choices}







\begin{figure}[t]
    \centering

    \includegraphics[width=0.49\linewidth]{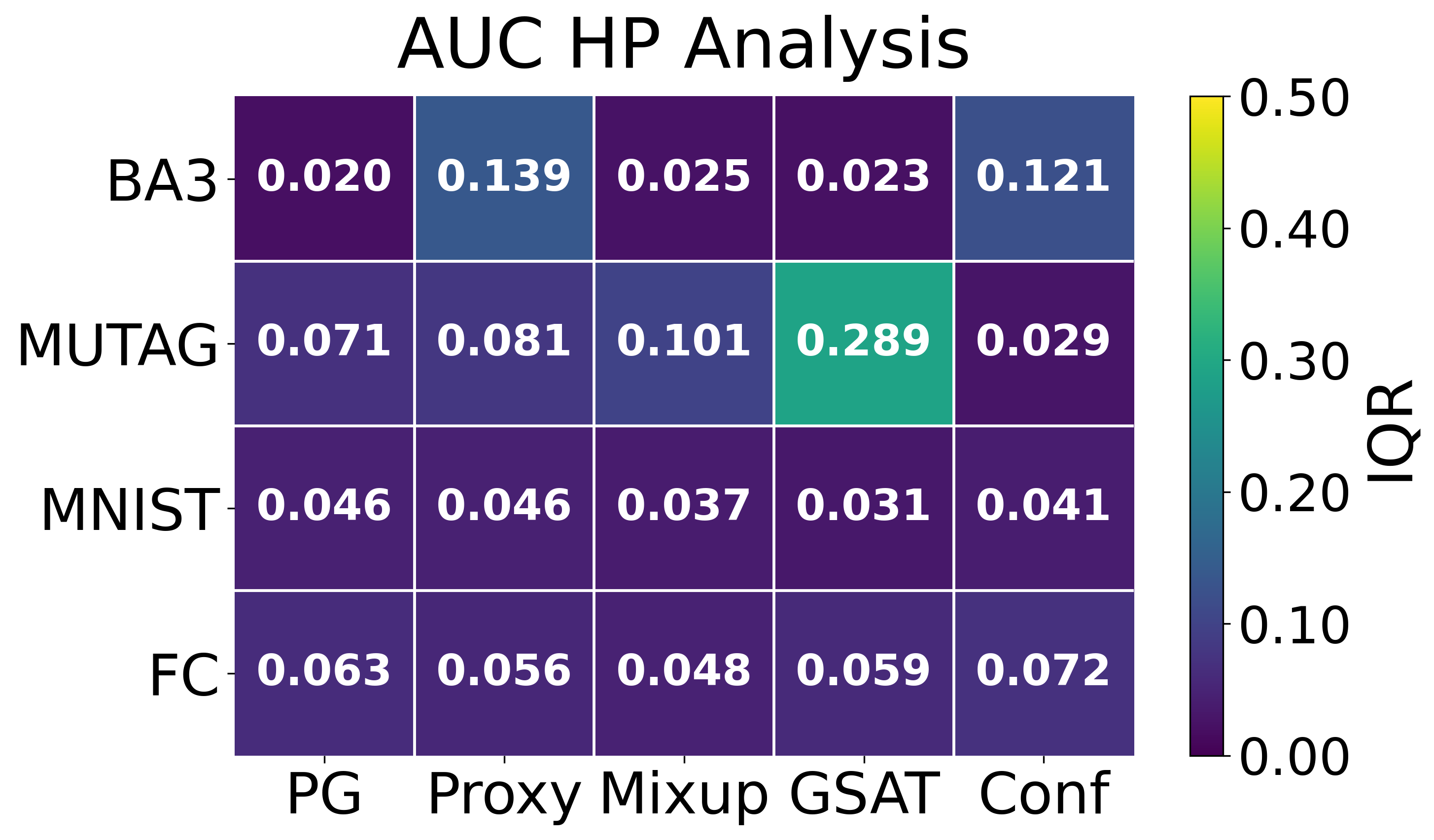}
    \hfill
    \includegraphics[width=0.49\linewidth]{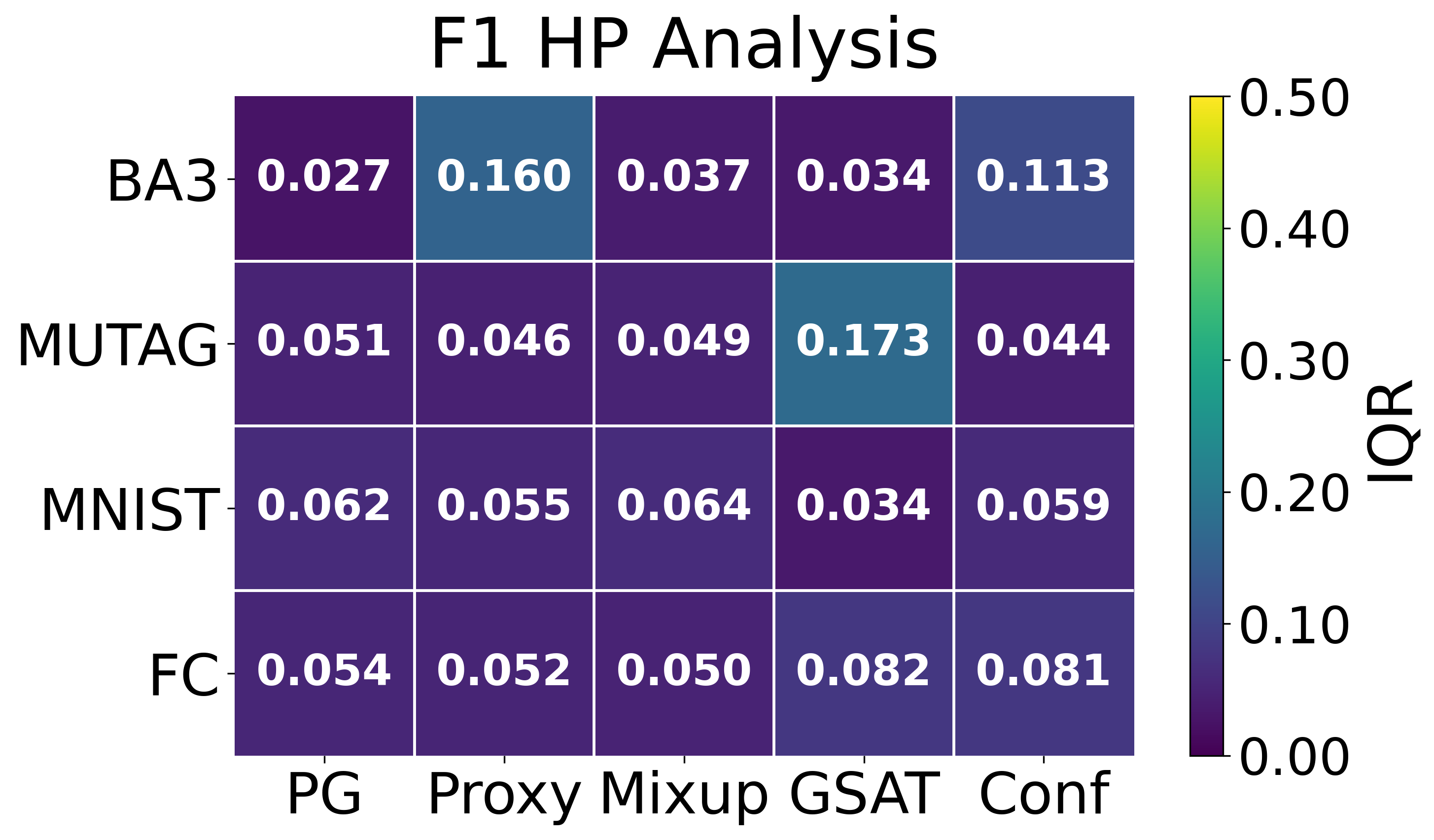}

    \includegraphics[width=0.49\linewidth]{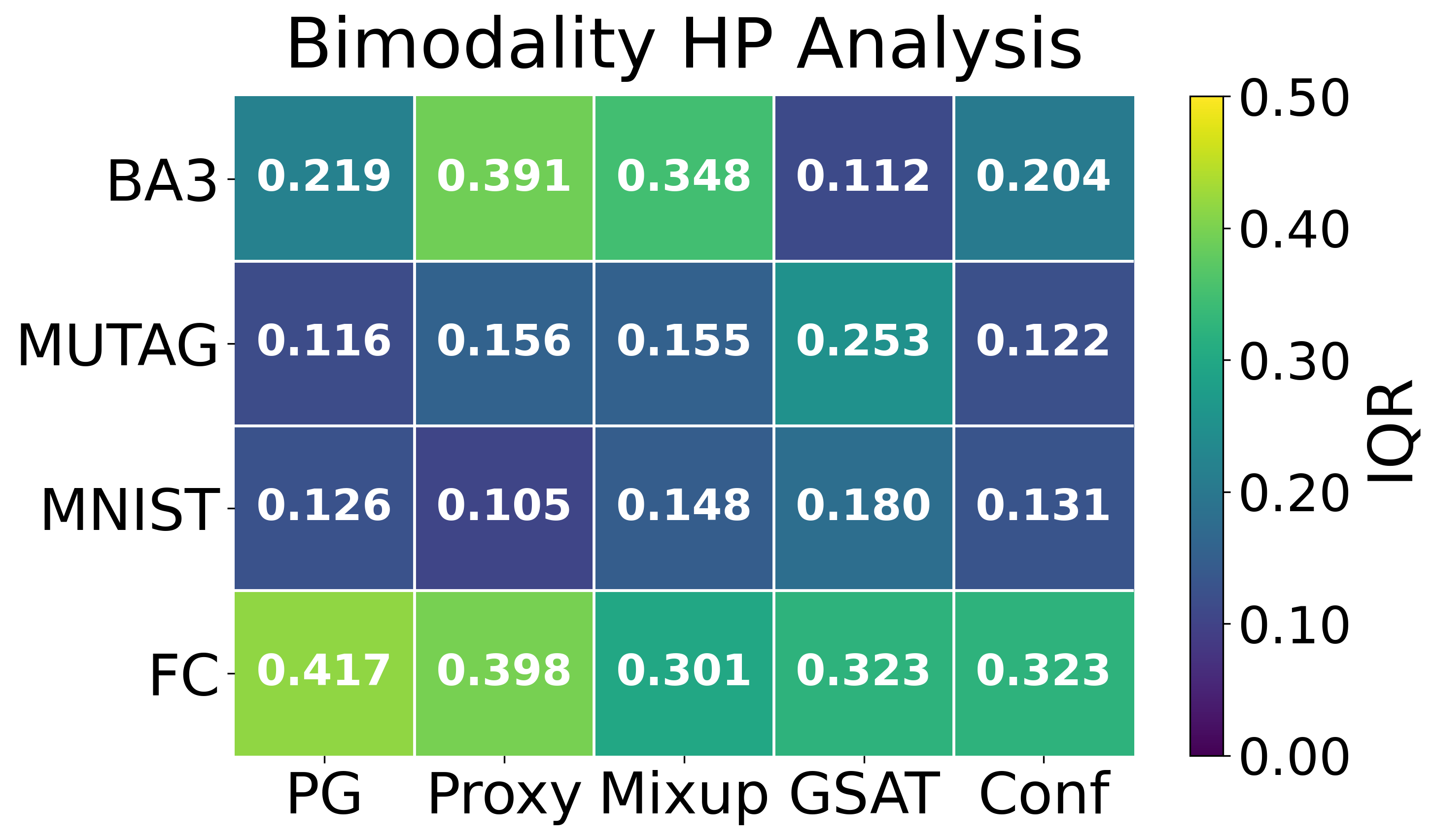}
    \hfill
    \includegraphics[width=0.49\linewidth]{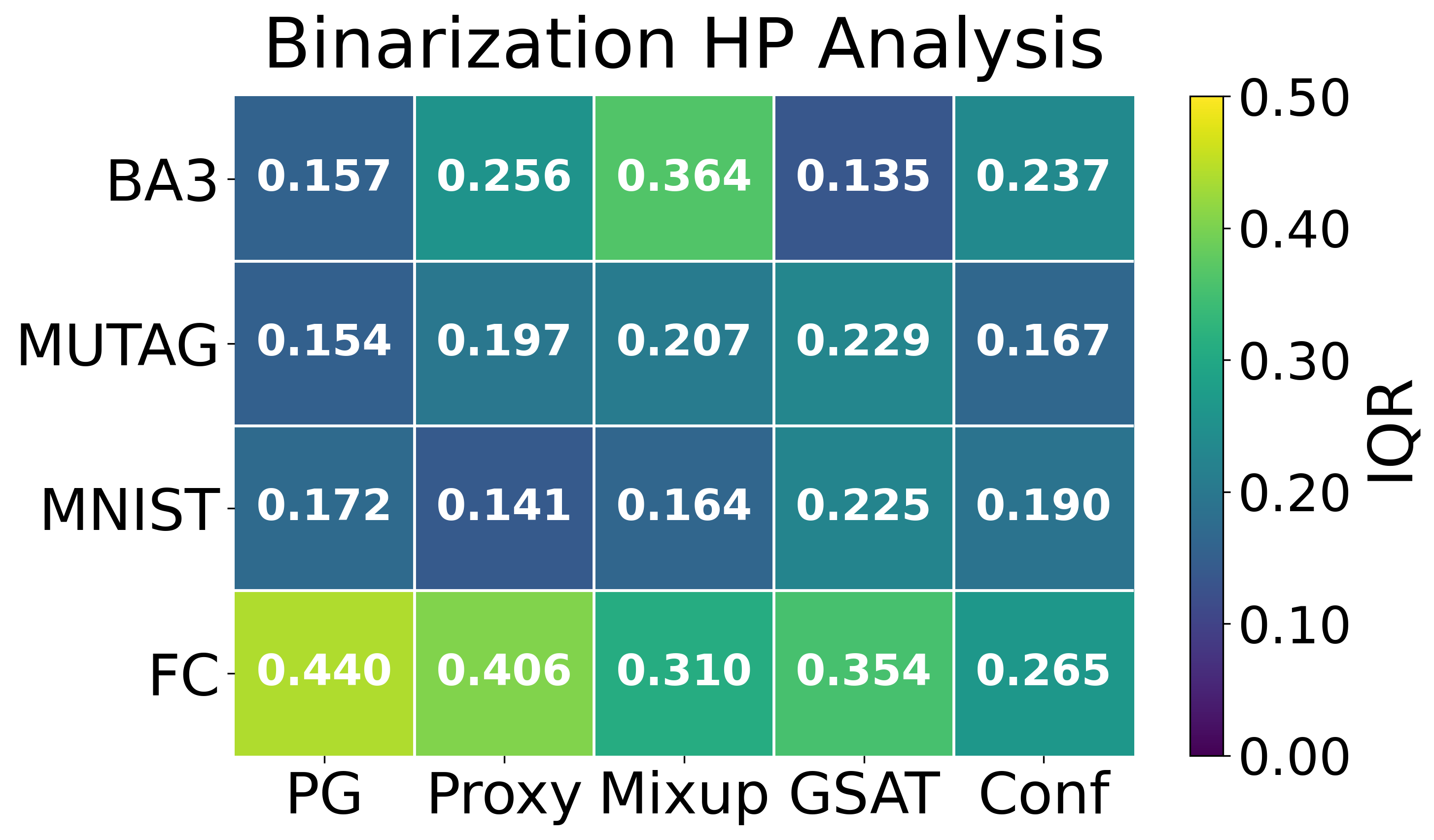}

    \caption{Hyperparameter sensitivity measured by the IQR of AUC, F1, bimodality, and binarization across combinations of $\alpha_0$, $c$, and $w^+$.
    The $x$- and $y$-axes denote base explainers and datasets, respectively.}
    \label{fig:hyperparam_analysis}
    \vspace{-15pt}
\end{figure}
Figure~\ref{fig:hyperparam_analysis} visualizes the hyperparameter sensitivity using heatmaps based on the interquartile range of the performance distributions.
Overall, the predictive metrics exhibit small IQR values across most datasets and base explainers.
Meanwhile, the distributional metrics tend to show slightly higher sensitivity, since they are directly affected by how aggressively the confidence zones polarize the score distributions.

\vspace{-5pt}
\subsection{Computational Cost}
\label{ssec:cost}



We measure the end-to-end training time of \OURS under the same experimental settings as the baselines, including validation and early stopping.
As shown in Table~\ref{table:time}, the additional cost of the multi-round refinement remains relatively small across datasets.
\OURS also exhibits efficient inference, introducing no noticeable overhead. 

\section{Conclusion}


We presented \OURS, which converts edge importance scores into pseudo-labels and refines them through self-guidance, producing more interpretable score distributions with clearer binarization and bimodality.
Our theoretical analysis explains binarization on pseudo-labeled edges and its conditional propagation to unlabeled edges.
Experiments on synthetic and real-world datasets show generally improved explanation accuracy across GIB and non-GIB base explainers.


\section*{Acknowledgments}
{
\small
This work was partly supported by the National Research Foundation of Korea (NRF) grants funded by the Korea government (MSIT) (RS-2024-00341425 and RS-2024-00406985),
the ``Advanced GPU Utilization Support Program'' funded by the Government of the Republic of Korea (Ministry of Science and ICT),
and the New Faculty Startup Fund from Seoul National University.
}


\bibliographystyle{IEEEtran}
\bibliography{reference}

%
%

\end{document}